\documentclass[10pt,twocolumn,letterpaper]{article}

\usepackage{cvpr}
\usepackage{times}
\usepackage{epsfig}
\usepackage{graphicx}
\usepackage{amsmath}
\usepackage{amssymb}
\usepackage{bm}
\usepackage{array}

\newcommand{\bs}[1]{\ensuremath{\boldsymbol{#1}}}

\cvprfinalcopy %

\def\cvprPaperID{2281} %

\ifcvprfinal\fi
\begin{document}

\title{A Pooling Approach to Modelling Spatial Relations for\\
Image Retrieval and Annotation}

\author{Mateusz Malinowski\\
Max Planck Institute for Informatics\\
Saarbr\"{u}cken, Germany\\
{\tt\small mmalinow@mpi-inf.mpg.de}
\and
Mario Fritz\\
Max Planck Institute for Informatics\\
Saarbr\"{u}cken, Germany\\
{\tt\small mfritz@mpi-inf.mpg.de}
}

\maketitle

\setlength{\textfloatsep}{7pt plus 1.0pt minus 1.0pt}

\begin{abstract}
Over the last two decades we have witnessed strong progress on modeling visual object classes, scenes and attributes that have significantly contributed to automated image understanding. On the other hand, surprisingly little progress has been made on incorporating a spatial representation and reasoning in the inference process. In this work, we propose a pooling interpretation of spatial relations and show how it improves image retrieval and annotations tasks involving spatial language. Due to the complexity of the spatial language, we argue for a learning-based approach that acquires a representation of spatial relations by learning parameters of the pooling operator. We show improvements on previous work on two datasets and two different tasks as well as provide additional insights on a new dataset 
with an explicit focus on spatial relations.
\end{abstract}

\section{Introduction}
\label{section:introduction}
In a daily life spatial concepts play an important role in human communication. Our comprehension and shared understanding of spatial concepts allow us to make references to specific objects as well as to resolve references made by others. The resolution of such references consists of two aspects, a linguistic part that expresses a relations and the involved concepts and perceptual part that allows us to perceive candidate entities that are involved in the mentioned relations. With spatial relations we can precisely localize object of our interest, ask an another person to act on that object, and expect from the person that first she understands the language of spatial relations and second she has a similar understanding of spatial relations in the environment. As we aim at building machines that ``understand'' and act upon our intention expressed in natural language, we need to also take care of learning spatial concepts from human data so that both -- machine and human -- refer to a common apprehension of spatial concepts that  are well aligned with each other.

Recent work that has addressed spatial language includes natural language commands for robotics  \cite{tellex2011understanding,guadarrama2013grounding} and question answering systems about the content of real-world scenes \cite{malinowski14nips} which relies on hand-crafted approach to spatial representations -- often driven by the need for high precision. However, it is also arguable beneficial for problems requiring high recall such as image search \cite{hodosh2013framing,lan2012image} where coverage on a wide range of spatial concepts becomes important. Yet we are missing techniques to automatically acquire and learn spatial relations to provide the desired coverage.
\begin{figure}[t]
\centerline{\includegraphics[width=1.0\linewidth]{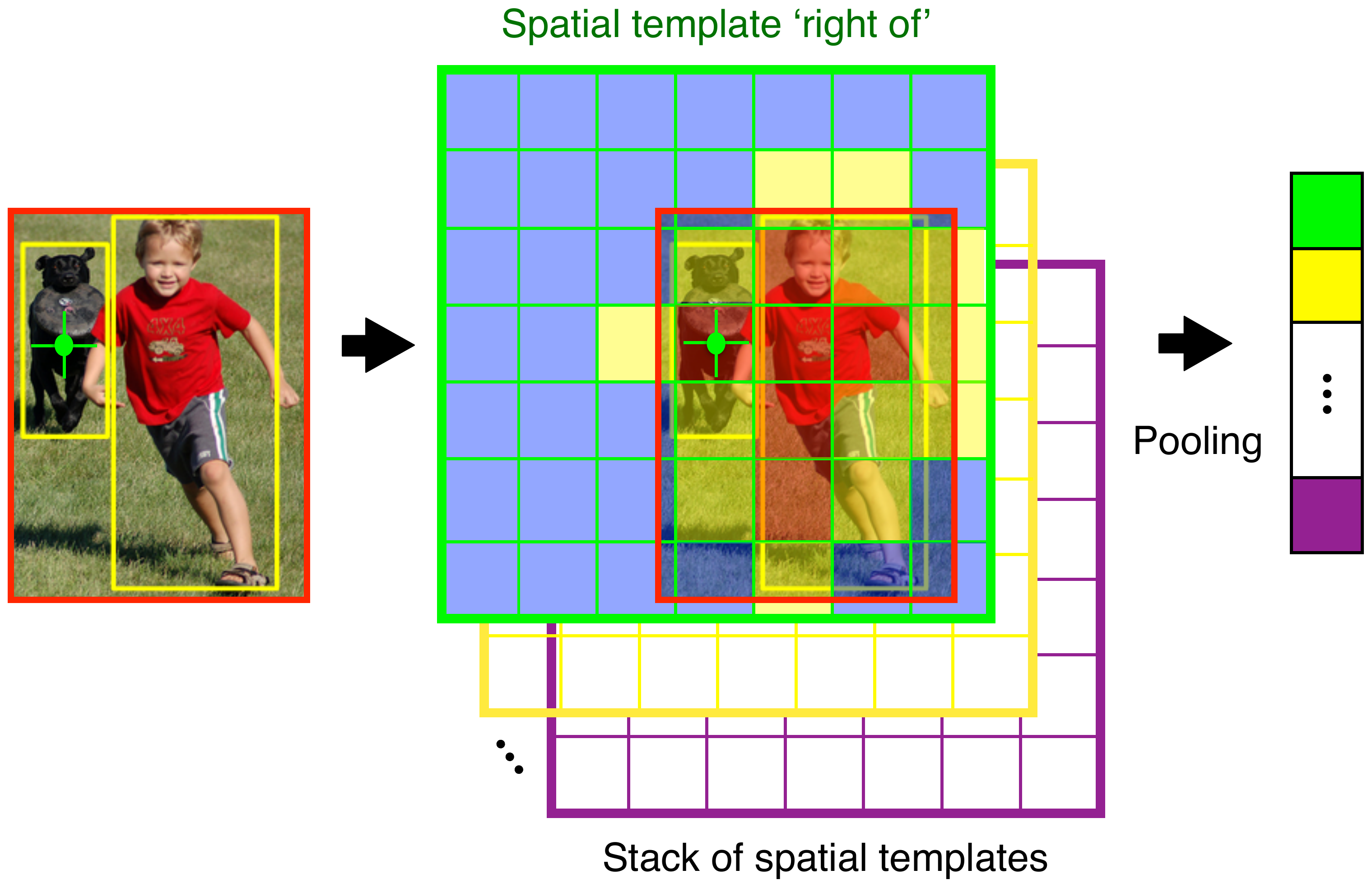}}
\caption{
We propose a pooling regions interpretation of deictic spatial relations, and show its importance for image retrieval and annotation tasks. We start from a spatial fragment representing a pair of detections: 'boy' and 'dog', and compute spatial representation by projecting the weighted pooling template at the center of the 'dog' detection and pooling the 'boy' localization accordingly.
}
\label{fig:movable_pooling}
\end{figure}

Apart from building spatial representations in machine perception, there is a long standing interest from psychologists in understanding how human apprehend spatial concepts \cite{logan1996computational,regier2001grounding}. Mainly based on differences in reference frames, they categorize spatial concepts into {\it basic}, {\it deictic} and {\it intrinsic} relations.
Moreover, the psychological studies also offer an interesting model of spatial relations, so called {\it spatial templates} \cite{logan1996computational}. 
In our work, we are interested in deriving representations of deictic spatial relations and their application to today's image retrieval and annotation methods. 
These relations express the position of one object with respect to other objects by projecting the observer's frame of reference onto the reference object, and can be modeled with spatial templates. 
Conceptually, a spatial template is associated with a spatial relation and represents regions of acceptability under the relation. It is centered at the object of reference and computes a goodness of the localization of another object with respect to the referent. 

In our work, we exploit that those models of spatial concepts are tightly related to the widely used pooling approaches in computer vision. We show in section \ref{section:learnable_spatial_templates}, spatial templates fit into a spatial pooling regions framework \cite{lazebnik2006beyond} by fusing ideas of learning pooling operators \cite{malinowski:2013:learnable_pooling} with object-centrism \cite{russakovsky2012object}.

Finally, we show that our approach to spatial reasoning readily extends two popular retrieval architectures \cite{lan2012image,karpathy2014deep} by showing a competitive or even improved results on a two datasets. We also further analyze 
our model on a new datasets 
with an explicit focus on spatial relations.
\\
{\bf Contributions}
In this work, we show how spatial pooling regions can be used for  spatial representations and reasoning by drawing a link between pooling operators and spatial templates. Next, we show that the spatial templates can be estimated from data if bounding boxes are available and there are spatial sentences of the form (object, spatial relation, object) associated with images. We estimate templates from two sources: our new data with human annotations, and data with automatically generated annotations according to some rules \cite{lan2012image} and point out differences in the obtained templates. The estimation procedures resembles the experimental scenarios in \cite{logan1996computational} but results are obtained from real-world images with many different object categories and implicit annotations of spatial arrangement. Finally, we extend two retrieval architecture \cite{lan2012image} and \cite{karpathy2014deep} to work with our spatial model. We show how an explicit representation of spatial relations improves performance quantitatively as well as qualitatively by showing the association between language and object on example images.

\section{Related work}
\label{section:related_work}
\noindent{{\bf Modeling spatial relations in images}}
Previous work has addressed the problem of image retrieval with structured object queries \cite{lan2012image} where the authors consider structured queries - a textual input with a binary spatial preposition between two nouns - together with a limited number of different spatial prepositions. %
Our work goes beyond structured queries and limited spatial vocabulary. For this purpose instead of using a hand-crafted representation of a set of only few relations ('above', 'below', and 'overlap' like in \cite{lan2012image}), we propose a flexible and learnable representation that is based on spatial templates \cite{logan1996computational}, and thus can be interpreted as a version of the learnable pooling regions \cite{malinowski:2013:learnable_pooling} centered at the reference object. 
\\
\noindent{{\bf Image-sentence alignment}}
While there have been successful methods that align sentences with images \cite{linvisual,kong2014you} the
recent research on embedding \cite{socher2013grounded,karpathy2014deep,mao2014explain} have opened a door for bi-directional methods that retrieve images based on a textual input, or sentences from a given image.
However, in contrast to our work, none of these methods use spatial reasoning to improve the alignment. \cite{karpathy2014deep} learns an embedding between textual and visual fragments, while other approaches between an image and a whole sentence. %
\\
\noindent{{\bf Spatial Pooling Regions}}
Spatial pooling has been proven to work well in many recognition tasks \cite{lazebnik2006beyond,yang2009linear} and is still a part of many recent approaches \cite{krizhevsky2012imagenet}. Although the research literature is densely populated with many variations of a spatial pooling regions framework, to the best of our knowledge there is no work that links pooling regions with spatial reasoning on object detections in a scene. In this work, we fill this gap and show a suitable interpretation of the framework.
Closely related to our work is an object-centric pooling \cite{russakovsky2012object} that relies on the object localization methods to distinguish between a foreground and background and next pool over both regions separately. 
Although, our method is also based on the localization of different objects, we spatially relate every pair of detections in the image to reason about their spatial arrangement.
\\
\noindent{{\bf Grounding Spatial Relations}}
Although research on  grounding of spatial language has a long standing tradition, previous methods mostly focus on rule-based spatial representation
\cite{moratz2006spatial,kruijff2007situated} or more recently on a set of hand-crafted spatial features with learnt weights
\cite{tellex2010,golland2010game,lan2012image,guadarrama2013grounding}. Although the latter approaches show  improvements they still rely on designing the right set of features and their generalization and scalability to many spatial relations have not been proven yet. \cite{lan2012image} uses only $2$ spatial prepositions, while \cite{golland2010game} and \cite{guadarrama2013grounding} concentrate on $11$. 

In our work, we propose a simple and uniform learning-based approach to spatial representation, and validate the proposed approach on different image-retrieval tasks with many spatial prepositions.

\section{Method}
\label{section:method}
We are proposing a representation for spatial relations and how it can be applied to image retrieval and annotation.
Motivated by the work on spatial templates \cite{logan1996computational}, we establish a connection between the popular pooling representations and the spatial templates.

First, we present our spatial model and describe how it is parameterized.
Then, we present an application of our approach 
to image retrieval setting \cite{lan2012image} with a restricted query language and where ground truth bounding boxes of different objects are available.
We proceed by showing how our spatial model can be incorporated into a fragment embeddings framework \cite{karpathy2014deep}. 
Here, annotated bounding boxes are unavailable and the query language is unrestricted.

Section \ref{section:learnable_spatial_templates} discusses a novel extension of a spatial pooling approach \cite{lazebnik2006beyond} to support spatial arrangement between detections.
In the following sections we show different instances of our model. Section \ref{section:estimating_spatial_distribution} discusses an application of the spatial templates where bounding boxes of different objects are known during the training and the query language has a restricted structure,  while section \ref{section:deep_fragments} shows how to extend the deep fragment embeddings \cite{karpathy2014deep} to work with spatial templates in unrestrictive setting without ground truth bounding boxes.

\subsection{Modeling spatial representations by spatial pooling}
\label{section:learnable_spatial_templates}

\noindent{{\bf Spatial basis}}
Spatial pooling framework \cite{lazebnik2006beyond} can be interpreted in terms of spatial basis 
\begin{align}
\label{eq:pooling_regions}
\Theta^k = \sum_{j=1}^M \bs{w}_j^k \circ \bs{u}_j
\end{align}
where $\bs{u}_j$ is an image feature located at position $j=(x,y)$ in the image, $\circ$ is a piece-wise multiplication, and $k$ refers to the k-th spatial pooling template. Hence, the standard spatial pooling with division into 2-by-1 subregions can be phrased in this representation as 
$\Theta^1 = \sum_{j=1}^{\frac{M}{2}} \bs{1} \circ \bs{u}_j + \sum_{j = \frac{M}{2} + 1}^M \bs{0} \circ \bs{u}_j$ and
$\Theta^2 = \sum_{j=1}^{\frac{M}{2}} \bs{0} \circ \bs{u}_j  + \sum_{j = \frac{M}{2} + 1}^M \bs{1} \circ \bs{u}_j$,
where $\left\{1, ..., \frac{M}{2}\right\}$ and $\left\{\frac{M}{2}+1, ..., M\right\}$ refer to the first and second half of the image respectively. 
Using such representation, the pooling operator can be included in a learning-based framework where the pooling weights $\{\bs{w}_j^k\}_{j,k}$ are jointly optimized together with a classifier \cite{malinowski:2013:learnable_pooling}. %
 Although, originally the logistic regression is used, the whole method is agnostic to the choice of a classifier and can be easily integrated with other objective functions with an additional hyper-parameter defining the size of the receptive field (or equivalently the discretization level) and the number of the pooling templates $\Theta^k$. 
\\
\noindent{{\bf The pooling interpretation of spatial relations}}
In psychology, \cite{logan1996computational} has proposed a theory of the spatial relations apprehension by estimating a fit of a spatial template. The template is centered at the reference object and models the relative locations of other objects in the environment.   Although the theory has existed for a long time in the psychological community, there is little work that includes similar concepts in modern computer vision architecture for a spatial reasoning. The theory identifies spatial templates with different spatial prepositions and represent those as score maps centered at the object of reference. The support of such score map covers the whole environment and it `softly' computes a spatial fit of a related object to the reference object under the relation by taking the score at the object's position. For instance, all the objects at the right position of the reference object gets a high score under the 'right template' and a low score under the 'left template'. Most strikingly, such templates can be interpreted in terms of the pooling regions with an image as the environment.

Consider a pair of detections representing 'dog' and 'boy' together with a statement 'A boy on the right side of a dog' as shown in Fig. \ref{fig:movable_pooling}. Let $x,y$ be the center of the 'dog' bounding box. Now, we place the center of the weighted spatial pooling regions $\Theta^{\text{right of}}$ at the position $x,y$ and pool over $N$x$M$ different subregions according to the weights. 
This produces a 
feature
that characterizes the fit of the localization of 'boy' according to the spatial template 'right of'. Here, $N$ and $M$ characterize the discretization level. Accordingly, our representation of spatial relations is computed as follows:
\begin{align}
\Theta^{\text{rel}}(i, \bs{u}^{(d)}) = \sum_{j=1}^M w_j^\text{rel} \circ u^{(d)}_{i-\frac{M}{2}+j} 
\end{align}
where $i$ is the position of the reference object, $\bs{u}^{(d)}$ is a score map representing the localization of the related object $d$ 
(e.g. a detector score map, or introduced in section \ref{section:deep_fragments} a Dirac image),
with value $0$ for positions outside of the image. In contrast to Eq. \ref{eq:pooling_regions}, $w_j^\text{rel}$ and $u^{(d)}_j$ are scalars. Latter represents the localization score map at position $j=(x,y)$.

In this work, we investigate two special cases of the more general spatial framework in the context of image retrieval.
First, in section \ref{section:estimating_spatial_distribution} we take advantage of the ground truth bounding boxes and initialize the pooling weights with the estimated spatial spatial templates (Fig. \ref{fig:relations_weights_viz}). In this scenario, we use queries with a limited structure and vocabulary.
Second, in section \ref{section:deep_fragments} we consider a challenging scenario with a complex natural language queries and where ground truth bounding boxes are missing. 

\subsection{Estimating spatial templates}
\label{section:estimated_spatial_distribution}
We consider a scenario with a restricted query language of the following form (noun, spatial preposition, noun) together with a limited vocabulary without inflection - for instance ('airplane', 'in front of', 'building'). Moreover, let assume the annotated bounding boxes are available during the training with the object categories from the same vocabulary. 
Thanks to those restrictions, and in contrast to section \ref{section:deep_fragments}, we can first estimate the spatial templates from data, and next initialize the pooling weights $\{w_j^k\}_{j,k}$ with the estimations.

To exemplify the estimation procedure, consider a spatial preposition 'above' and take all the images that are annotated with a sentence containing 'above', for instance ('picture','above','bed'). Next, we center a spatial template representing 'above' at the center of 'bed' bounding box and copy the content of the 'picture' bounding box. Afterwards, we proceed to the subsequent image with 'above' annotation and repeat the 'copying' procedure while storing the already copied contents. To obtain smooth spatial templates, we create the localization score map by filling the whole 'picture' bounding box with ones and take it as its content. Finally, we use such derived spatial template as the initialization of $\{w_j^\text{above}\}_j$. Figure \ref{fig:relations_weights_viz} shows the estimated spatial templates for spatial relations that we use in our experiments. Since the initialization already acts as a strong regularization, unlike in section \ref{section:deep_fragments}, we do not need to resort to discretization of the image space into large subregions - in other words we consider one pixel sized receptive fields. Note that our estimation is still based solely on the descriptions of the image and does not require directly annotating spatial relations.

In section \ref{section:experiments} we visually inspect the estimated templates. Interestingly, the estimation procedure and our visualizations resemble the experimental scenarios in \cite{logan1996computational} where templates are estimated from the points drawn by humans on a frame with respect to a given spatial preposition. Our case is however different in that we collate results based on real world images with many object categories and implicit spatial arrangement. That is, for every sentence of the form (object, spatial relation, object), participants of the experiment only annotated which images satisfy the sentence.

Next section shows how to include a spatial model into the state-of-the-art method on a retrieval task with missing ground truth bounding boxes and unconstrained language.

\label{section:estimating_spatial_distribution}

\subsection{Deep fragment embeddings with spatial reasoning}
\label{section:deep_fragments}
\noindent{{\bf Deep fragment embeddings}}
The main goal of \cite{karpathy2014deep} is to retrieve relevant images based on a sentence query, and conversely. 
The model learns a bi-directional embedding on a set of unconstrained images and corresponding sentences.
As opposed to previous work on embedding, it finds a mapping between visual fragments represented as the image-induced activations of the bottleneck layer of the most certain detections \cite{girshick2014rcnn}, and textual fragments that are represented as triplets of the form $(R,\bs{t}_1,\bs{t}_2)$, where $\bs{t}_1$ and $\bs{t}_2$ are 1-of-k word encodings under a binary dependency relation $R$ \cite{dependency_parser}. Moreover, the framework does not require any annotated associations between the textual and visual fragments nor even annotated bounding boxes. Instead, it incorporates a MIL \cite{chen2006miles} procedure into the learning process. The objective function consists of two parts: global ranking objective that learns the image-sentence similarities that are consistent with the ground truth annotations, and fragment alignment objective that is based on the intuition that for a given textual fragment at least one of the bounding boxes in the corresponding image should have a high score with this fragment.
The learning process optimizes a linear combination of both objectives and aims at finding a good inner-product based similarity between the fragments.
For a detailed exposition of the objective function, we refer the reader to \cite{karpathy2014deep}.

We use both, textual $\bs{s} = f\left(W_R \begin{bmatrix} W_e \bs{t}_1 ;  W_e \bs{t}_2 \end{bmatrix} + b_R \right)$, and visual fragments $\bs{v} = W_m \left[ \text{CNN}(I_b) \right]$ in our work.
Here, $W_e$ is a fixed $400,000 \times 200$ matrix that encodes a 1-of-k vector into a $200$-dimensional distributed representation \cite{huang2012improving}, $f$ is RELU activation function \cite{glorot2011deep}, and $\text{CNN}(I_b)$ is a $4096$ dimensional activations of the bottleneck layer induced from the image fragment $I_b$. The fragment embedding weights $W_R$, $W_m$, and $b_R$ are learnt jointly using the aforementioned objective function so that the score $\bs{v}^T \bs{s}$ is high for the fragments that match well, low otherwise.
\\
\noindent{{\bf Spatial extension}}
In addition to the visual and textual fragments, we introduce spatial fragments that are based on the pooling interpretation of spatial relations. Let $\Theta^k(O_j, \cdot)$ be a 
weighted spatial division that represents k-th spatial concept centered at the position of j-th detection. Here, $O_j = (x_j,y_j)$ represents the center of the j-th bounding box.
We can formally cast such representation into the spatial pooling framework as follows. Let $\bs{u}^{d}$  be a Dirac image associated with detection $d$. It is $u^{d}_{(x,y)} = 1$ if $(x,y)$ is the center of the bounding box $d$ and $u^{d}_{(x,y)}=0$ at other positions. For every pair of detections, we consider the reference detection $d_1$ and build a Dirac image $\bs{u}^{d_2}$ of the related detection. Next, we place the spatial template $k$ at $O_{d_1}$ - the center of the reference detection - and pool over the Dirac image $\bs{u}^{d_2}$, producing a spatial fragment $\Theta^k(O_{d_1}, \bs{u}^{(d_2)})$.
We repeat such procedure for every pair of detections, with the 1st and 2nd elements of the pair as the reference and related detections, finally producing a $D^2$  such spatial fragments for every spatial concept, where $D$ is the number of detections. 

Such representation can be transformed into the matrix-vector multiplication framework, which is consistent with \cite{karpathy2014deep}, by pulling out the weights and a discretization of the image space: $\bs{p} = W_s g(\bs{u}^d)$. Here, $\bs{u}^d$ is the Dirac image of a detection $d$, $g(\bs{u})$ takes a Dirac image $\bs{u}$, discretize it into $N$-by-$M$ subregions, and subsequently vectorize it. The matrix $W_s$ is a mapping from $N M$ dimensional vector space into a $K$ dimensional space of  spatial concepts. Note that, although this space can directly correspond to $K$ different prepositions, it can also be treated more abstractly with $K$ chosen based on a validation set.

Analogously, we define spatio-textual fragments
\begin{align}
\bs{z} = f\left(W_z \begin{bmatrix} W_e \bs{t}_1 ;  W_e \bs{t}_2 \end{bmatrix} + b_z \right)
\end{align}
 where $W_z$ maps from the $400$ dimensional representation of both words into a $K$ dimensional space of spatial concept. Finally, we use the same objective function to train the weights so that $\bs{p}^T \bs{z}$ give a high score for the matching spatial fragments and a low score otherwise.

\section{Experiments}
\label{section:experiments}

\begin{table*}[t]
\centering
\scalebox{0.95}{
\begin{tabular}{|ccccccccccc|cc|}\hline

\multicolumn{13}{|c|}{Estimated spatial templates}\\\hline

\multicolumn{11}{|c|}{Our extended human queries} & \multicolumn{2}{c|}{Structured \cite{lan2012image}} \\\hline
 
\multicolumn{11}{|c|}{} & \multicolumn{1}{c}{} \vspace{-0.3cm} & \multicolumn{1}{c|}{} \\

\includegraphics[width=0.053\linewidth]{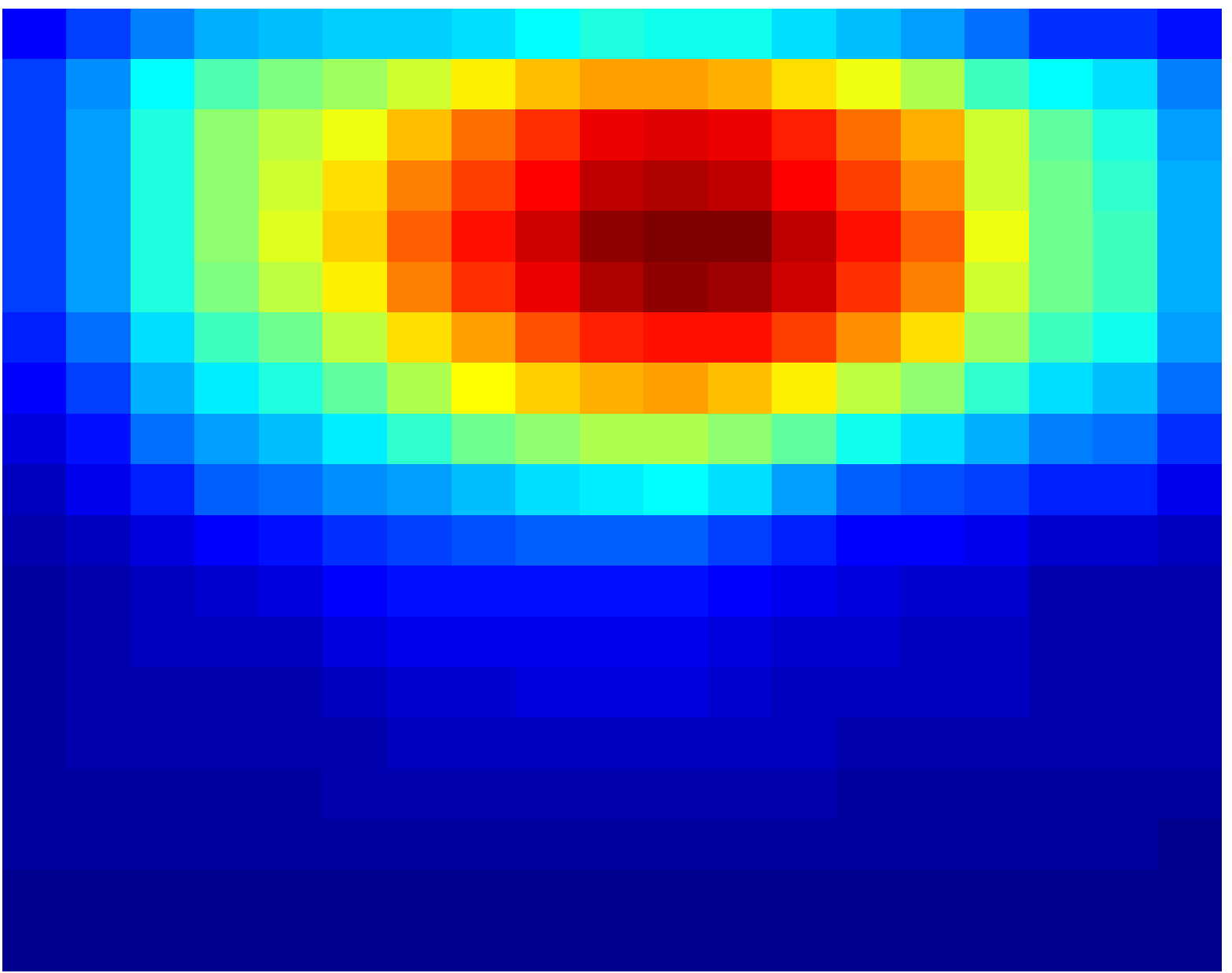} &
\includegraphics[width=0.053\linewidth]{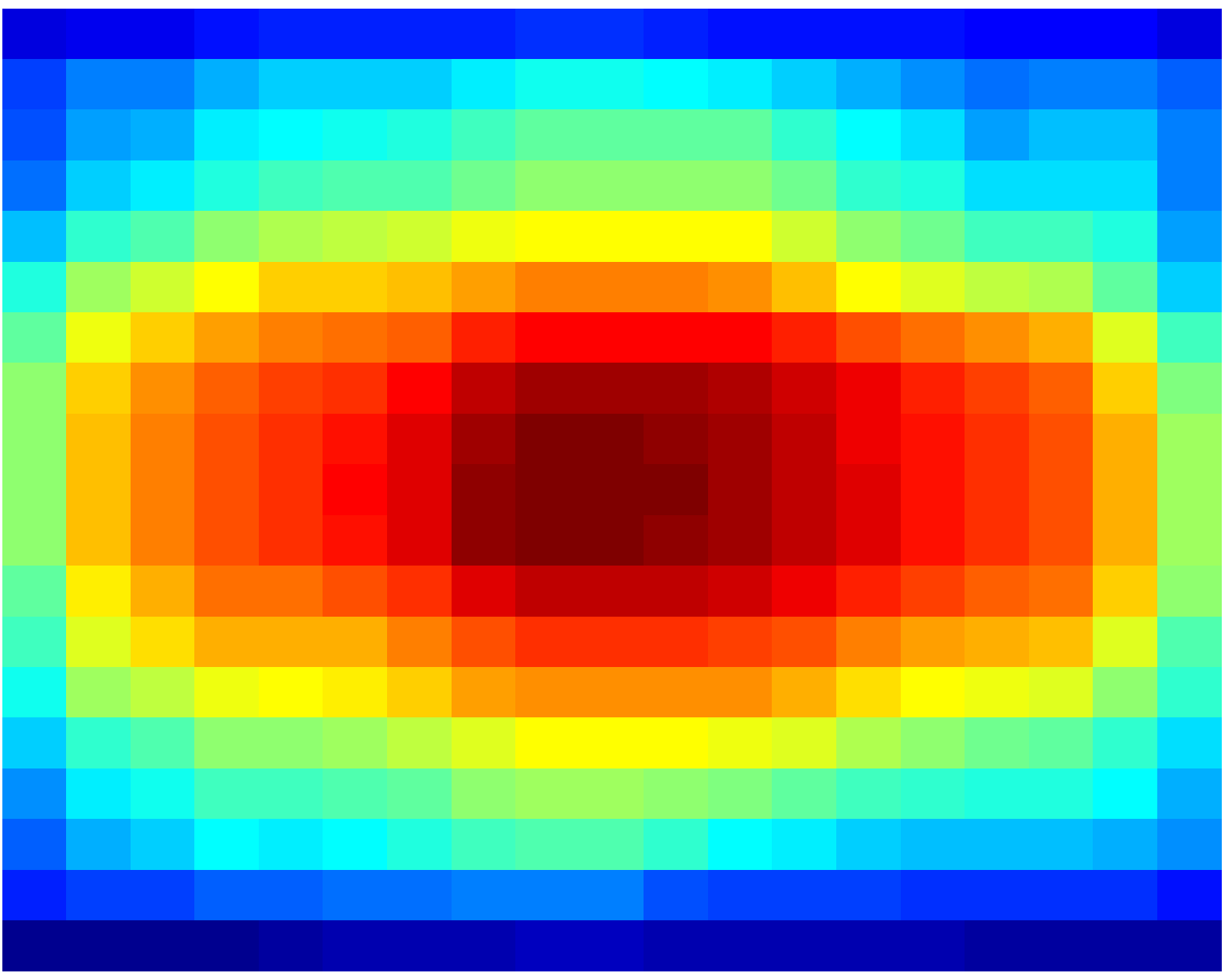} &
\includegraphics[width=0.053\linewidth]{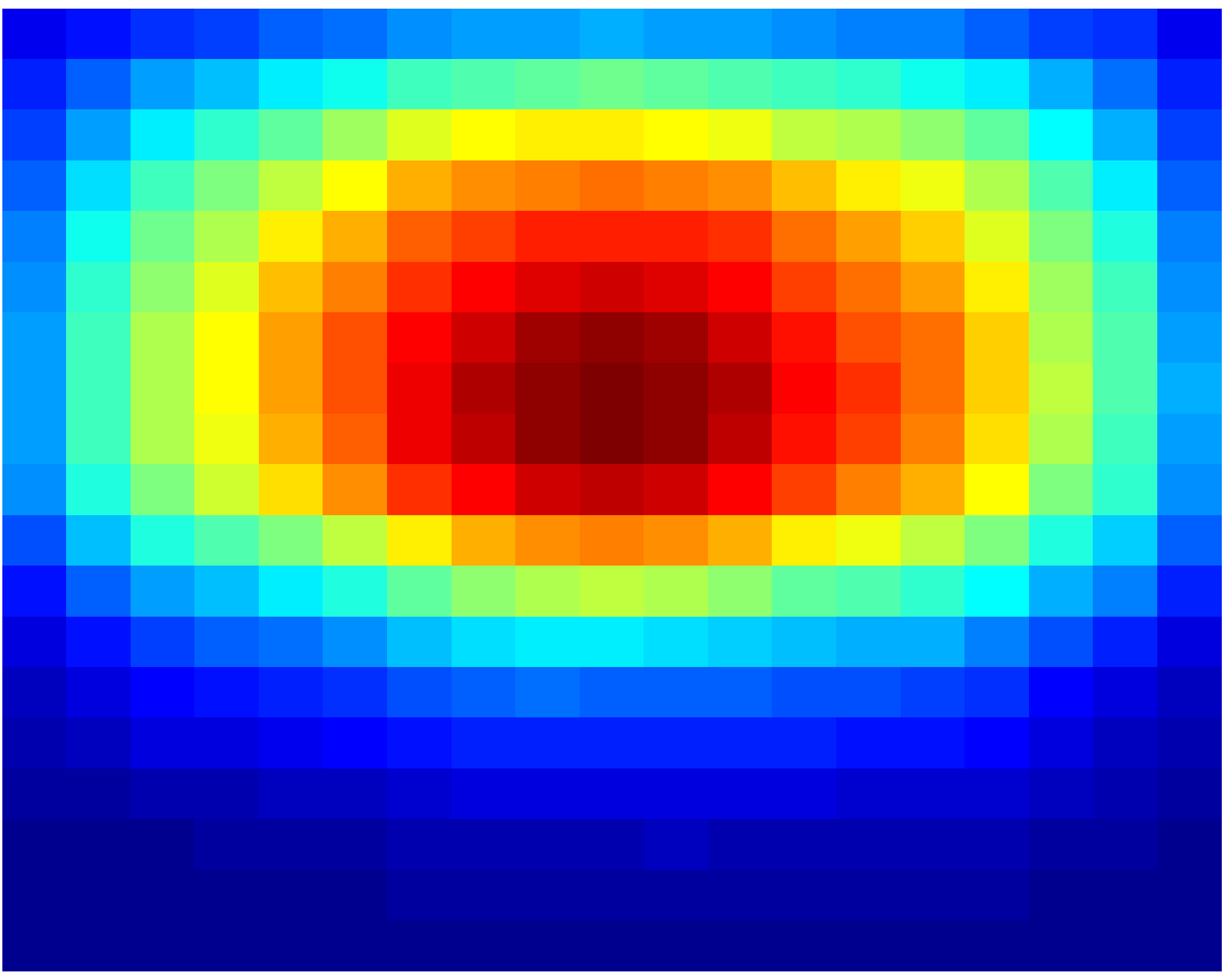} &
\includegraphics[width=0.053\linewidth]{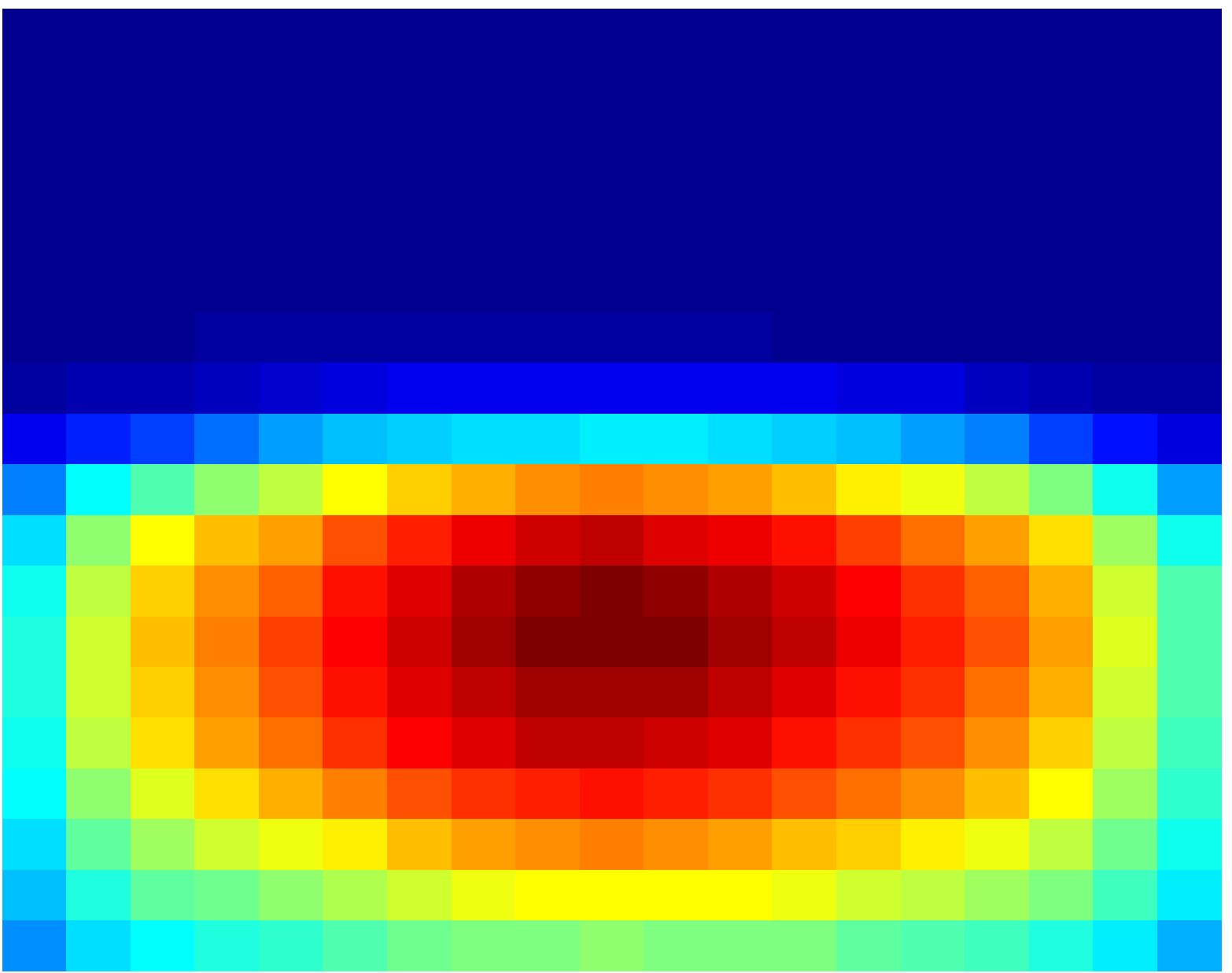} &
\includegraphics[width=0.053\linewidth]{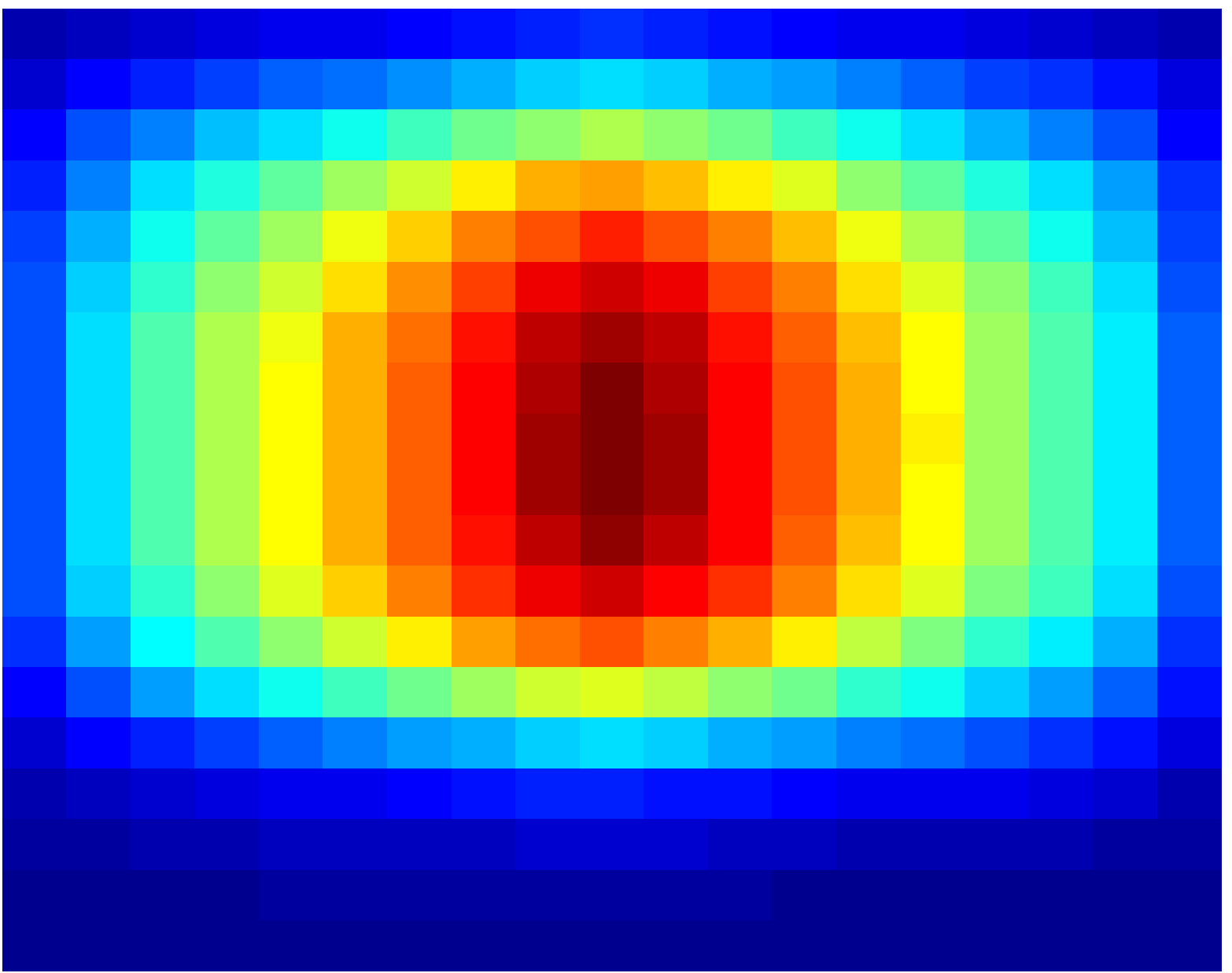} &
\includegraphics[width=0.053\linewidth]{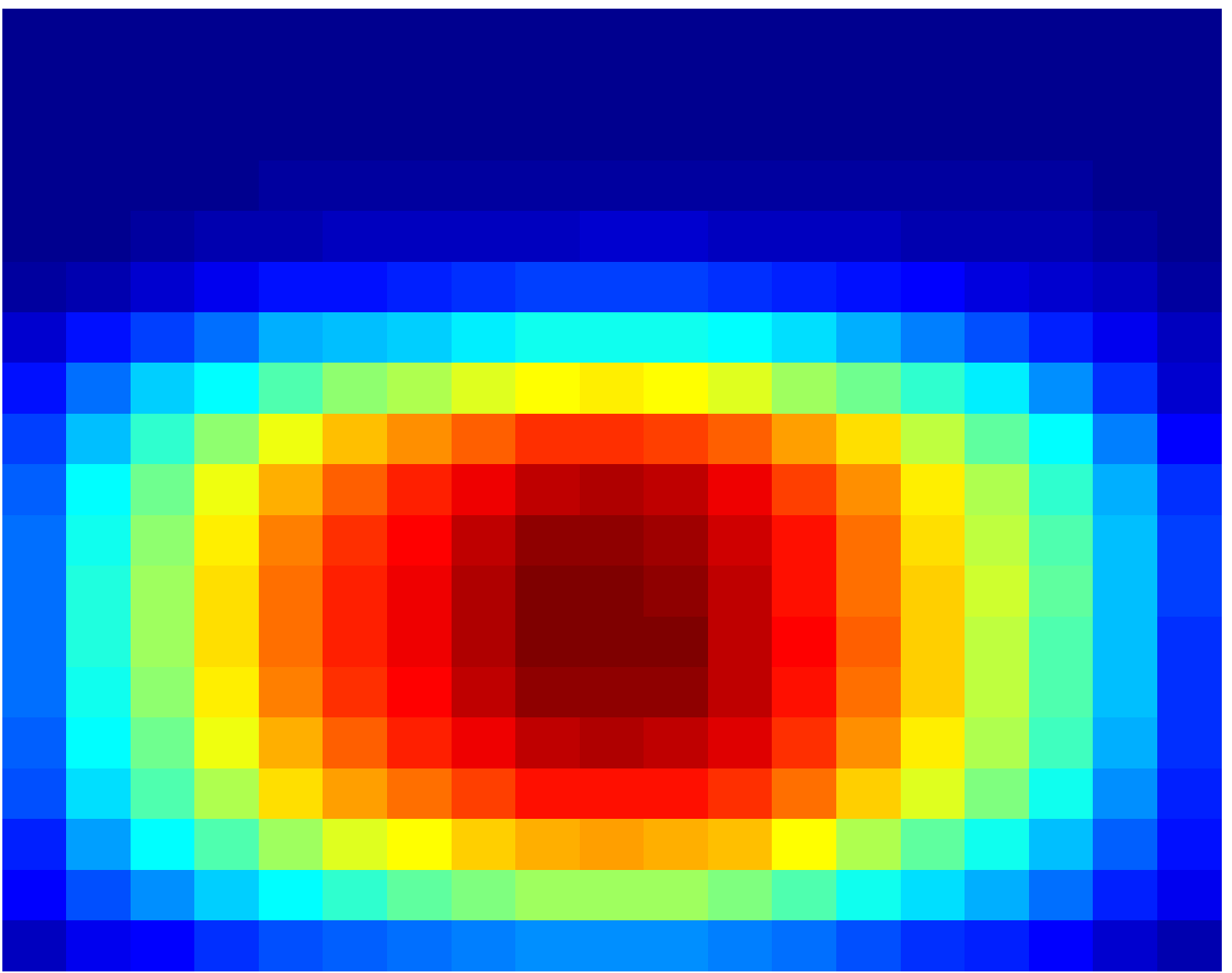} &
\includegraphics[width=0.053\linewidth]{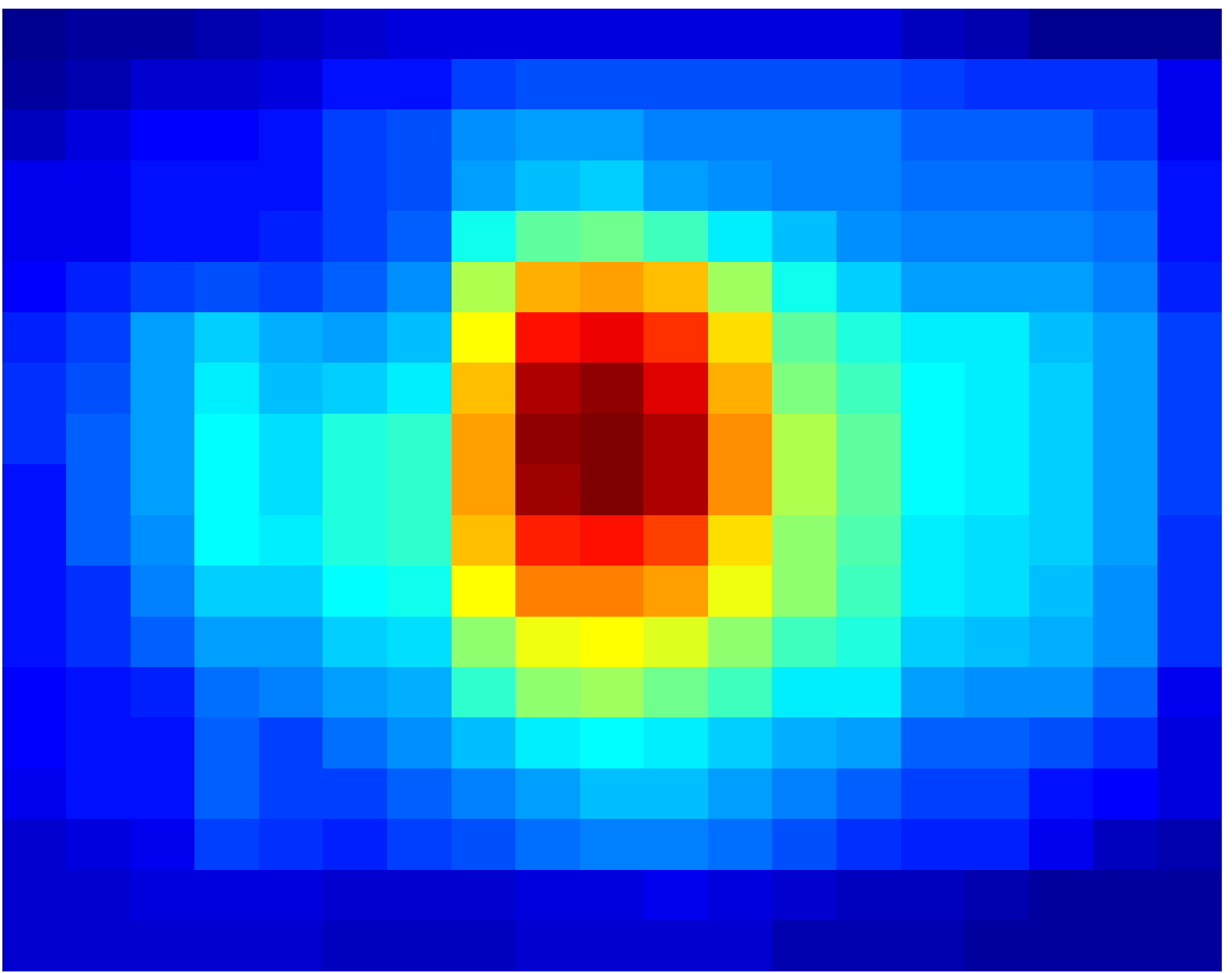} &
\includegraphics[width=0.053\linewidth]{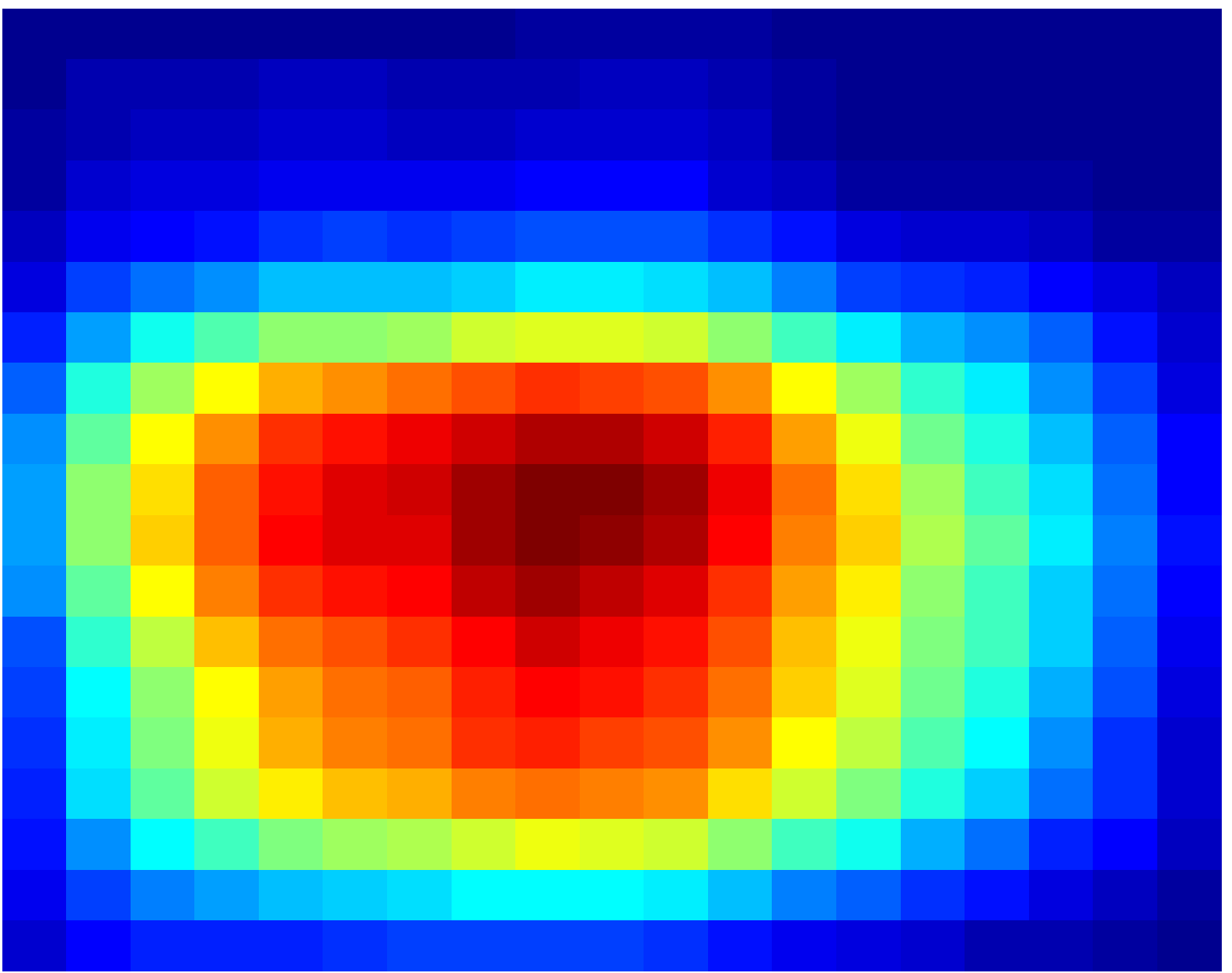} &
\includegraphics[width=0.053\linewidth]{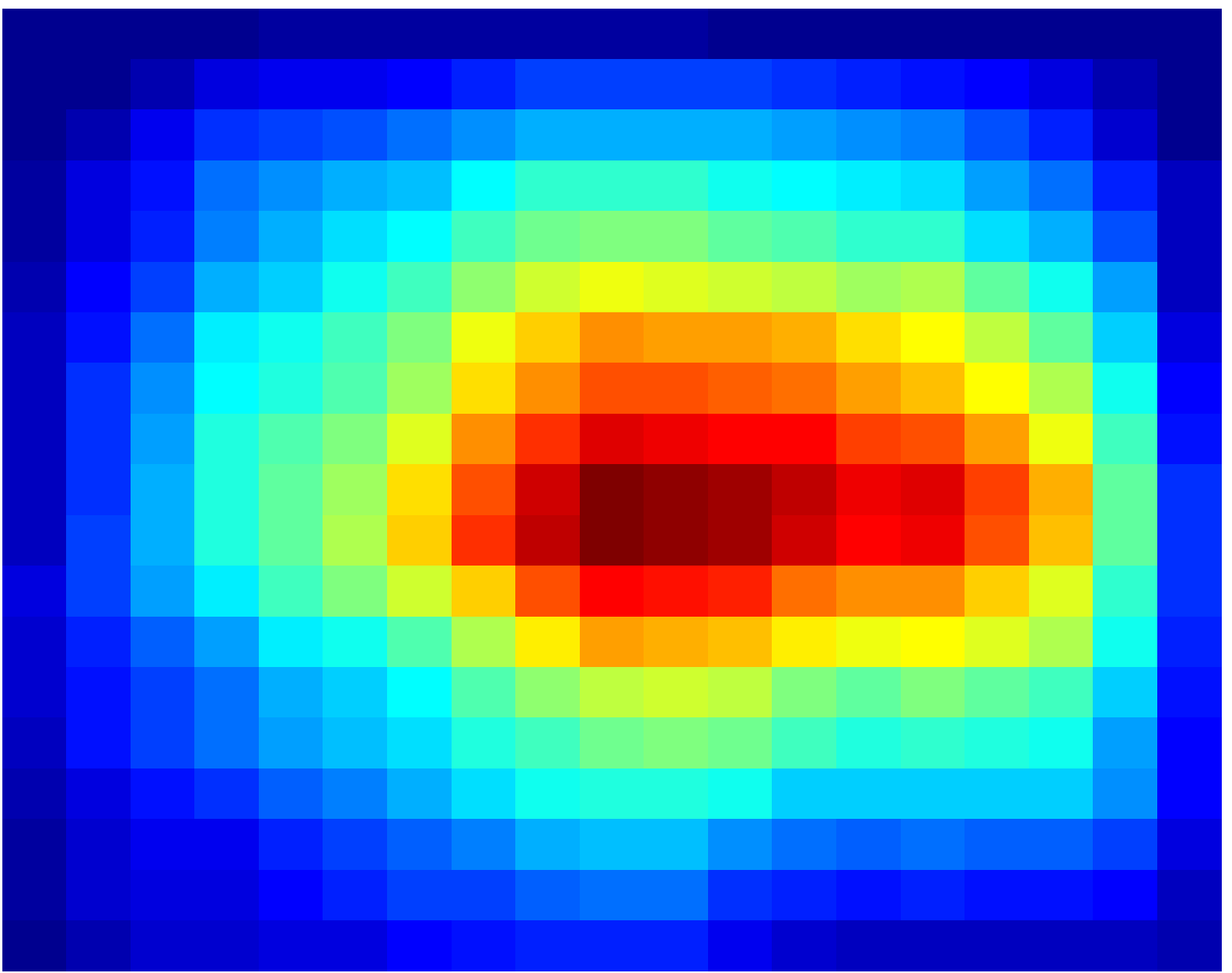} &
\includegraphics[width=0.053\linewidth]{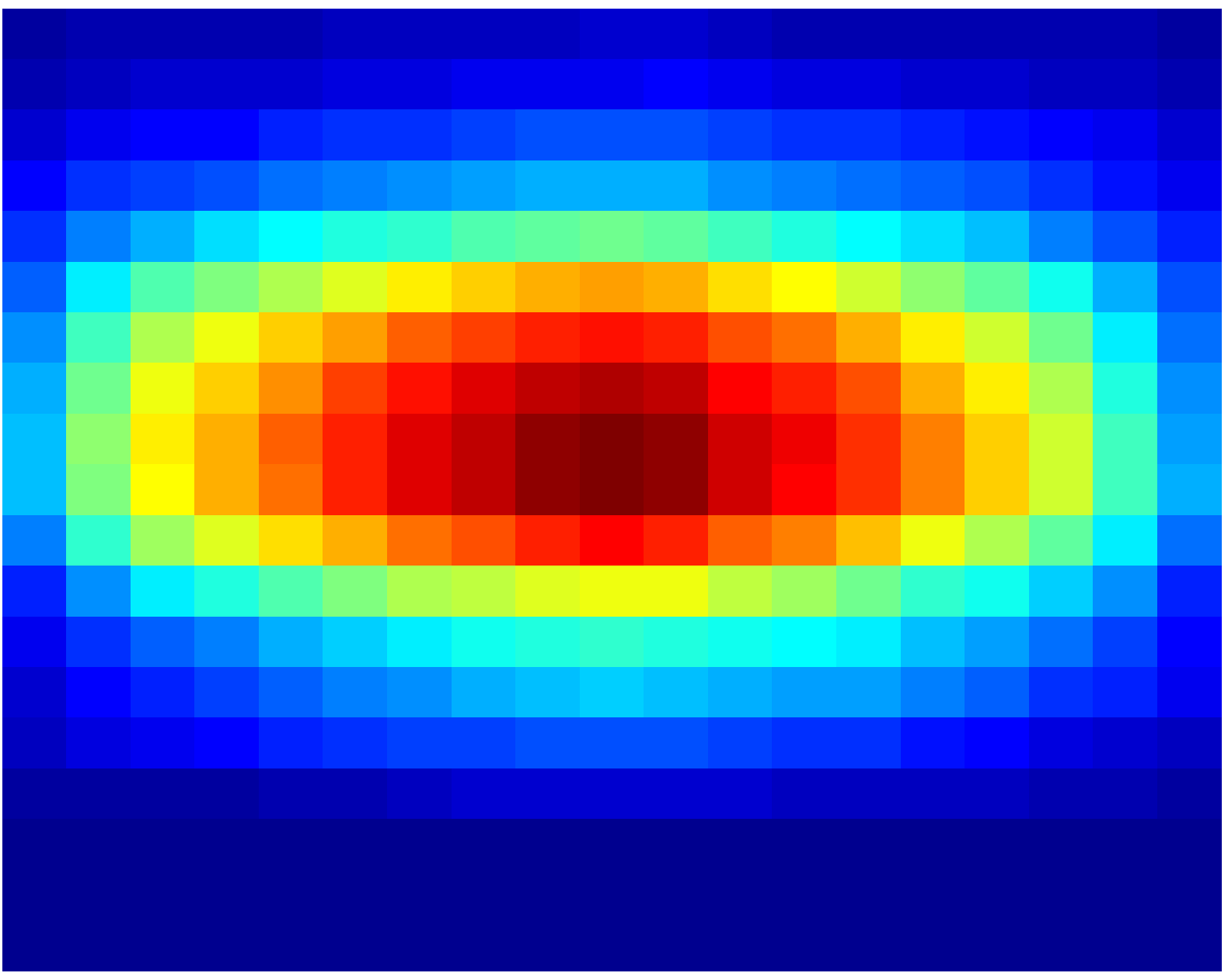} &
\includegraphics[width=0.053\linewidth]{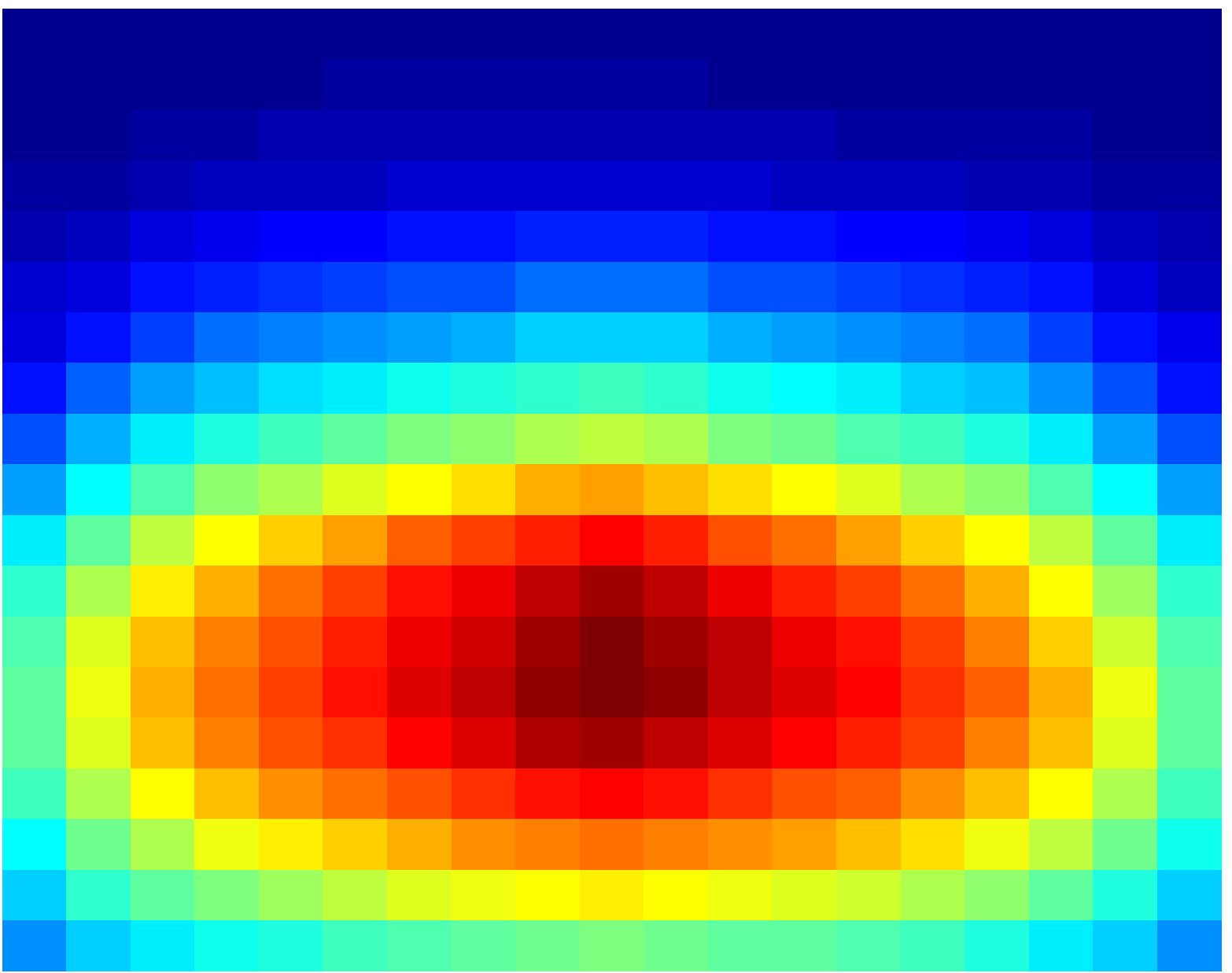} &
\includegraphics[width=0.053\linewidth]{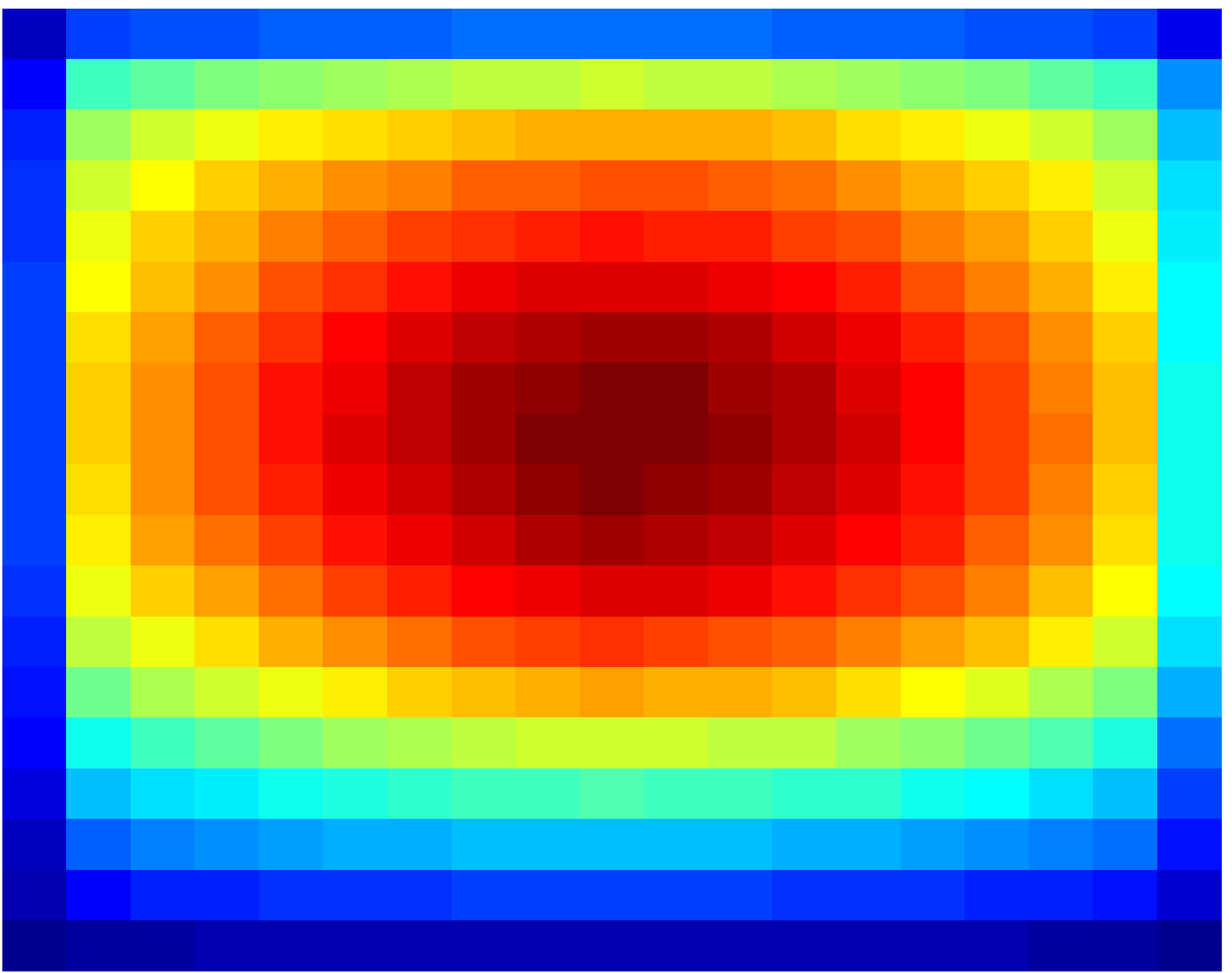} &
\includegraphics[width=0.053\linewidth]{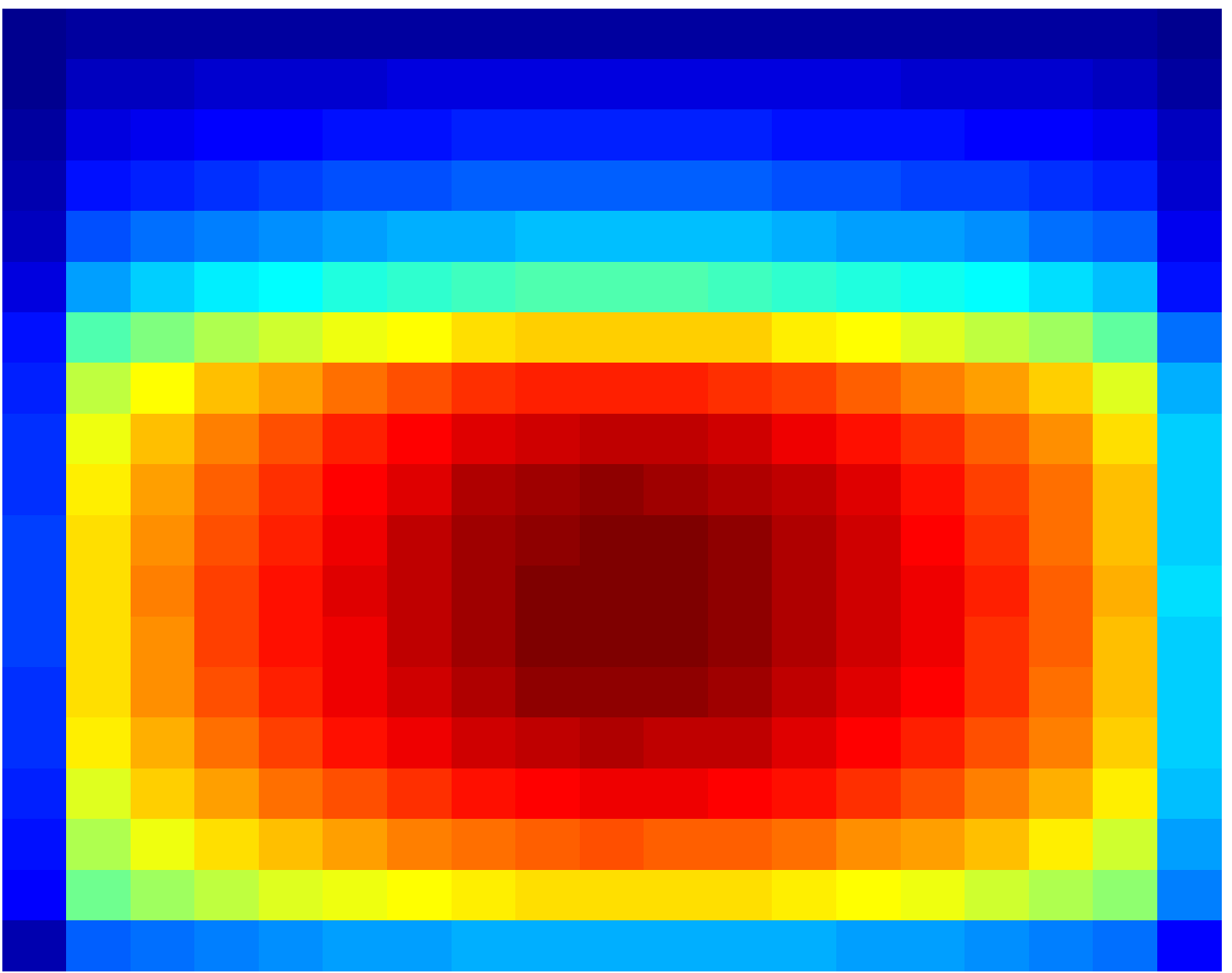}

\\\hline
above & across from & behind & below & in & in front & inside & left & right & on & under & above & below 
\\\hline

\end{tabular}
}
\caption{Visualization of estimated spatial templates.
The red color encodes high values and blue color low values.
}
\label{fig:relations_weights_viz}
\end{table*}

We conduct experiments on several datasets. First, two retrieval datasets use a constrained query language that allow us to use annotated bounding boxes during the training. 
Here, we estimate spatial templates as described in section \ref{section:estimated_spatial_distribution}. Both datasets augment the SUN09 image dataset with queries. The first dataset is introduced by \cite{lan2012image} and uses automatically generated queries, while the second dataset is our extension of \cite{lan2012image} with a human annotated queries and a wider range of spatial relations. Note that the difference between both annotation procedures is substantial, as in our dataset we deal with human notion of spatial concepts that are inherently ambiguous. In addition to the queries, both datasets include annotations which images are relevant to a given query. Again, our proposed annotations are based on human judgement. The last and the most challenging dataset, Pascal1k \cite{rashtchian2010collecting}, is a collection of images with associated natural language sentences. Although it does not contain the relevance annotations, it can still be used for a retrieval task \cite{socher2013grounded,karpathy2014deep}.

\subsection{Dataset}
\noindent{\bf Images}
All our experiments are based on real-world images. The SUN09 dataset \cite{choi_cvpr10} consists of $12,000$ annotated images with more than $200$ object categories. We use $4367$ images for training and $4317$ images for testing - the same split as in \cite{choi_cvpr10} and \cite{lan2012image}. The second dataset consists of $1000$ PASCAL images \cite{everingham2008pascal,rashtchian2010collecting}. Here, we follow \cite{karpathy2014deep} and use $800$ images for training, $100$ for validation, and $100$ for test.\\
{\bf Evaluation measures}
To be consistent and comparable with \cite{lan2012image} we use Mean Average Precision (mAP) across all queries to measure the performance of different methods on our first two datasets. This measure favors the retrievals with high precisions.
Similarly, for the sake of consistency with \cite{karpathy2014deep}, we use Recall@k (R@k) and Mean Rank (mean r) performance measures \cite{hodosh2013framing}. Recall@k computes the fraction of times the correct result is found among the top k retrievals. This measure favors high recall retrievals and is motivated by the search engines where it is more important to retrieve correct retrievals among top k results.
\\
{\bf Structured queries}
Structured queries are introduced in \cite{lan2012image}, but were not formally defined. Here, we formalize the notion of structured queries. 
We say that a query $q$ is structured if it has the form: $q := q_1 \wedge q_2 \wedge ... \wedge q_n $, where $q_i$ denotes either a noun or a triplet (noun, preposition, noun).
\\
{\bf Our dataset of structured queries with richer and human-based spatial language}
We use the structured queries from \cite{lan2012image} of the form (noun, spatial preposition, noun) with spatial prepositions such as 'above' and 'below', and extend such set to have queries with more spatial prepositions: 'left of', 'right of', 'in front of', 'behind', 'inside of', 'on', 'under', 'across from' and 'in'. 
We collect annotations by first asking in-house annotators to describe randomly selected images from the SUN09 dataset. Only tuples of the form ('noun', 'spatial preposition', 'noun') are permitted. In the second pass we curate this dataset and arrive at $53$ structured queries. 
Finally, the annotators annotate a binary relevance of each image according to every query. Since the latter requires a lot of human effort we have automatized the process by showing only images containing all objects described in a query. In this process, we have collected about $450,000$ relevance annotations and $53$ structured queries. In both passes, we instruct the annotators to take an observer's frame of reference.
Although our dataset uses a more restrictive query language than \cite{rashtchian2010collecting}, it is still challenging due to the use of human notion of spatial relations and high variations of object appearance in real-world images.
Although, ideally we would annotate also all spatial relations in every image, this process turns out to be too expensive as it scales up quadratically wrt. the number of objects in the scene per relation. Therefore, we decide on a more scalable approach where only descriptions of the relations are given. 

Compared with \cite{lan2012image}, our dataset consists of more spatial prepositions. In additions, our annotations are generated by  human annotators while the previous dataset uses a hand-crafted spatial model that is used to generate image descriptions as well as in the inference. %

Compared with \cite{rashtchian2010collecting}, our dataset provides a more reliable comparison with ground truth for the image retrieval task due to our relevance annotations. In addition, instead of focusing on the all aspects of the language, it is mostly about spatial relations.%

\subsection{Evaluation}

We investigate several experimental scenarios. First, we compare our method against previous work on the structured queries \cite{lan2012image}, where we show that with learnt spatial templates we can achieve comparable results to hand-crafted representations of spatial relations,
but under much weaker assumptions. 
Second, we also establish a baseline on our new dataset with human-based spatial relations and show that our method can learn an extended set of spatial concepts.
Third, we show the benefits of using spatial relation during the inference on a complex task with unconstrained natural language queries and real-world images without exploiting ground truth bounding boxes \cite{rashtchian2010collecting,karpathy2014deep}. 
Fourth, we visually investigate the estimated templates, and show improvement in alignment between language fragments and images.

\paragraph{Comparison to previous work on structured queries}
In order to establish a comparison to previous work on structured queries, we run experiments on the structured queries from \cite{lan2012image} and compare to their approach in Table \ref{table:baseline}. This dataset consists of $862$ ($463$ for training and $399$ for testing) queries of the form (noun, preposition, noun) with $111$ nouns. Their experiment contains only two different spatial relations: `above' and `below'. 
In this dataset, the spatial relations are automatically extracted by a hand-crafted formula on the $(x,y)$ coordinates of bounding boxes and serve as exact definitions of the spatial relations.
This spatial model is also used by the system of \cite{lan2012image} during the inference. In contrast, we assume that the procedure of generating queries is unknown to our system and we aim at obtaining good representations of the spatial relations only from data.
The  model of \cite{lan2012image} implements a structured SVM approach and models both the spatial relationship between objects in the query and co-occurrence between non-query and query objects via the compatibility function:
\begin{align}
	\label{eq:equation_from_tian}
    \sum_{i \in V_{q}} \alpha_i^T f(I(l_i)) + \sum_{i \in V_{q}} \sum_{j \in \mathcal{X} \setminus V_{q} } \gamma_{ij}^T f(I(l_j)) \\
      + \sum_{i,j,k \in E_{\mathcal{Q}}} \beta_{ijk} d_{\mathcal{Q}}(l_i,l_j,k) \nonumber
\end{align}
Here, $\alpha$, $\gamma$ and $\beta$ are weights learnt by the classifier, $V_{q}$ is a set of all objects (nouns) in the query, $\mathcal{X}$ is a set of all objects available during training, $f(I(l_i))$ is a HOG descriptor extracted \cite{lsvm-pami} at location $l_i$, $E_{\mathcal{Q}}$ denotes a set of object pairs and their spatial relations present in the query $\mathcal{Q}$, and $d_{\mathcal{Q}}(I_i,I_j,R_k)$ is used spatial model between detections $I_i$ and $I_j$ under the spatial relation $R_k$.
The last term is equal to $1$ if detections $l_i$ and $l_j$ are consistent under the spatial relation $k$, and is equal to $0$ otherwise. The consistency is determined via the same set of rules that are used to create queries.
This method achieves a performance of $11.16\%$ mAP without global features on queries of the type (noun, spatial preposition, noun).
Moreover, we also report the results of two more baselines (special cases of Eq. \ref{eq:equation_from_tian}): Part based detector where the sum of maximum response scores from each object detector is used as a score and the MARR model \cite{lan2012image}. The latter uses object detections as the features for the classifier and models co-occurrence between the detections (the second term in Eq. \ref{eq:equation_from_tian}), but without a spatial model. 

Since we are mostly interested in learning spatial relations, we implement the same compatibility function (Eq. \ref{eq:equation_from_tian}) but with our spatial component $\Theta^k(O_2, \cdot)$ that represents a spatial filter representing preposition $k$ and centered at the localization of the detection with a category pointed by a query (object $1$, preposition, object $2$). This matching between the category names and queries is possible since both use the same vocabulary for the objects.  For the same reason, we use 'preposition' to index different spatial templates, hence $K$ is equal to the size of spatial vocabulary. Note that here, we compute a spatial relationship between a pair of detections with categories extracted from the query. As Table \ref{table:baseline} shows, our approach achieves comparable results  at $11.12\%$ to the state-of-the-art despite the fact that we did not assume knowledge on the underlying representation of the spatial relations that the data was generated with. The first two rightmost entries in Table \ref{fig:relations_weights_viz} show the templates that we have estimated from data to capture a notion of the spatial relations. 

\begin{table}[t]
\centering
\scalebox{1.0} {
\begin{tabular}{| l | c |}
  \hline
  \multicolumn{2}{|c|}{{\bf Structured queries}}\\\hline
	\hline
	 Method & mAP  \\
	\hline 
 	Part based detector \cite{lsvm-pami} & $7.76\%$ \\
	MARR \cite{marr}  & $10.01\%$  \\
  Structure model \cite{lan2012image}  & $11.16\%$ \\
	\hline
  Our model & $11.12\%$ \\\hline
  \hline
  \multicolumn{2}{|c|}{{\bf Extended dataset of human queries}}\\\hline
  Our model & $7.90\%$ \\
  \hline
\end{tabular}
}
\caption{
Performance of our model that uses estimated spatial templates to other baseline approaches.
Note that Structure model uses the same rules to generate questions with spatial prepositions and during the inference.
}
\label{table:baseline}
\end{table}

\begin{figure}[b]
\centerline{\includegraphics[width=1.0\linewidth]{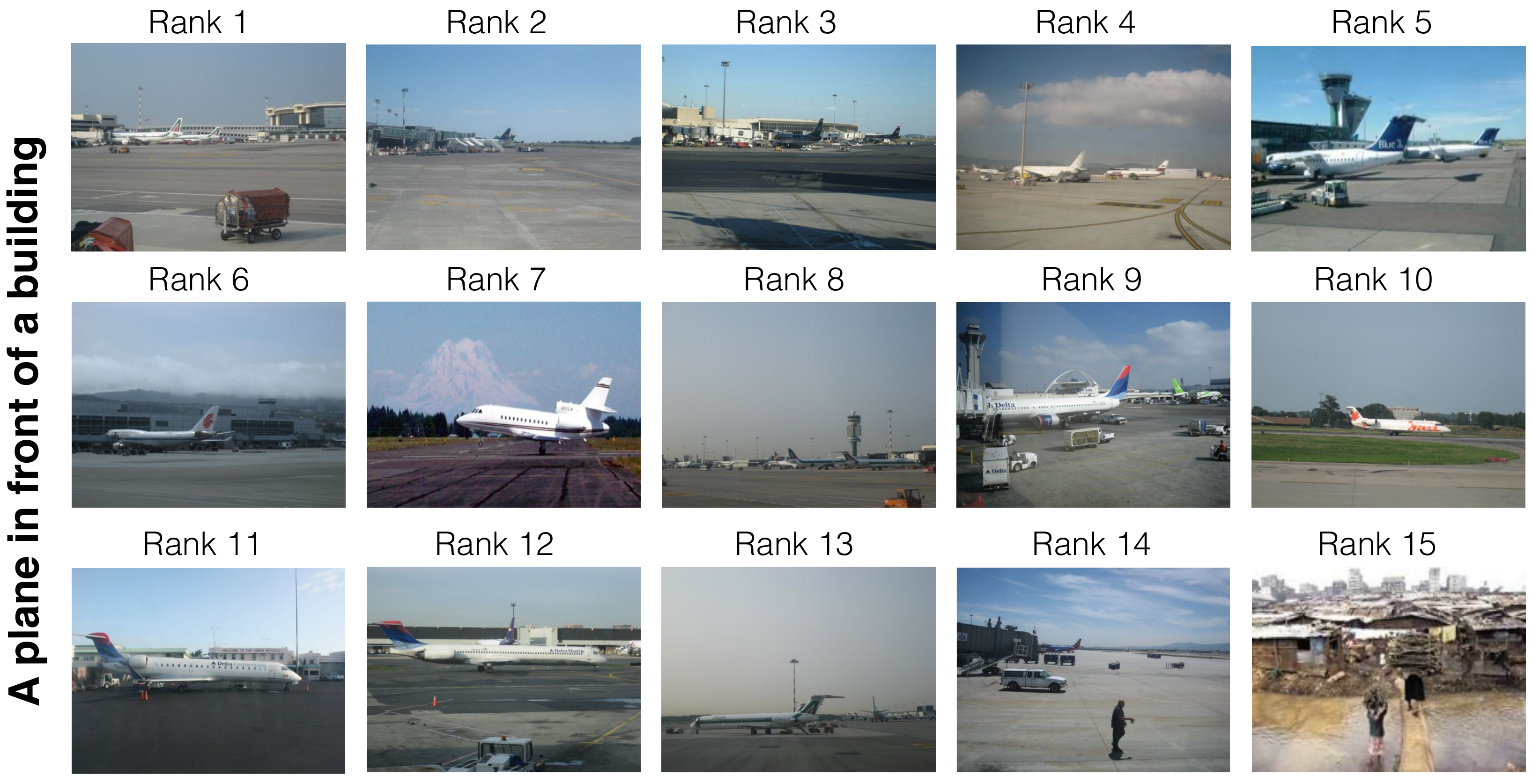}}
\caption{
Top ranked retrieved images from the query 'An airplane in front of a building' (SUN09 image dataset and our set of human queries). We see a high recall achieved by our method and two clear mistakes - Rank 7 and Rank 15.
}
\label{fig:plane_in_front_of_building}
\end{figure}

\paragraph{Extended set of spatial relations with queries annotated by humans}
We extend our analysis to our new dataset that contains an extended set of spatial relations that are -- in contrast to the previous dataset --  collected from human annotations. Since the exact human notion of spatial concepts is unknown, it has to be acquired from data. The second part of Table \ref{table:baseline} (Extended dataset of human queries) shows the performance of our approach, which achieves $7.90\%$ mAP, on our collected data with human queries. Note a drop in performance compared to the previous experiments as this is a more challenging setting. 
\\
{\bf Visualization of spatial templates}
To gain more insights about the spatial concepts apprehension, we visualize the estimated templates.
The first $11$ leftmost entries in Table \ref{fig:relations_weights_viz} show the spatial templates estimated on our new dataset. They follow our intuitions about the spatial layout (e.g. 'in' and 'inside' templates are much more focused than other spatial templates). More importantly, our visualization suggests that human apprehension of 'above' and 'below'  relations clearly differ from the procedure used to generate queries in \cite{lan2012image}, both are more focused in our case.
Interestingly, even if 'below' and 'under' are synonyms, the corresponding templates are not exactly the same. This suggests a slightly different human apprehension of both concepts. Also, pairs 'left'/'right' and 'above'/'below' are not entirely symmetrical. Although, some concepts such as 'in front' or 'behind' are rather three dimensional, it is still interesting to see how humans perceive them in a plain image. 
\\
{\bf Analysis of retrieved images}
We show the retrieved images by our architecture given an example query ('plane', 'in front of', 'building'). Figure \ref{fig:plane_in_front_of_building} shows the images together with their corresponding ranks. Further analysis revealed that most mistakes come from failure modes of the object detectors that our and \cite{lan2012image}'s methods are based on. Although there are stronger object detectors \cite{girshick2014rcnn} than part based models \cite{lsvm-pami}, we decide to keep the latter for the sake of consistency with \cite{lan2012image} and since our work is mainly concerned about spatial concepts.

\paragraph{Experiments on Pascal1k with unconstrained queries}
\begin{table}[t]
\centering
\scalebox{1.0} {
\begin{tabular}{| l | c | c | c | c|}
  \hline
  \multicolumn{5}{|c|}{{\bf Pascal1k}}\\\hline
  \multicolumn{5}{|c|}{Image Retrieval} \\\hline
  Method & R@1 & R@5 & R@10 & Mean r \\\hline
  Random Ranking \cite{karpathy2014deep} & $1.6$ & $5.2$ & $10.6$ & $50.0$ \\ 
  Socher et al. \cite{socher2013grounded} & $16.4$ & $46.6$ & $65.6$ & $12.5$ \\
  kCCA \cite{socher2013grounded} & $16.4$ & $41.4$ & $58.9$ & $15.9$ \\
  DeViSE \cite{frome2013devise} & $21.6$ & $54.6$ & $72.4$ & $9.5$ \\
  SDT-RNN \cite{socher2013grounded} & $25.4$ & $65.2$ & $84.4$ & $7.0$ \\
  Deep Fragment \cite{karpathy2014deep}  & $25.0$ & \boldsymbol{$69.4$} & $83.8$ & $6.9$ \\\hline
  Our model & \boldsymbol{$29.0$} & $68.6$ & \boldsymbol{$85.2$} & \boldsymbol{$6.7$} \\
  \hline\hline
  \multicolumn{5}{|c|}{Image Annotation} \\\hline
   Method & R@1 & R@5 & R@10 & Mean r \\\hline
  Random Ranking \cite{karpathy2014deep} & $4.0$ & $9.0$ & $12.0$ & $71.0$ \\ 
  Socher et al. \cite{socher2013grounded} & $23.0$ & $45.0$ & $63.0$ & $16.9$ \\
  kCCA \cite{socher2013grounded} & $21.0$ & $47.0$ & $61.0$ & $18.0$ \\
  DeViSE \cite{frome2013devise} & $17.0$ & $57.0$ & $68.0$ & $11.9$ \\
  SDT-RNN \cite{socher2013grounded} & $25.0$ & $56.0$ & $70.0$ & $13.4$ \\
  Deep Fragment \cite{karpathy2014deep} & $37.0$ & $69.0$ & $84.0$ & $10.4$ \\\hline
  Our model & \boldsymbol{$38.0$} & \boldsymbol{$70.0$} & \boldsymbol{$86.0$ } & \boldsymbol{$10.3$} \\
  \hline
\end{tabular}
}
\caption{
Performance of our model that uses a learnable spatial pooling framework to learn the spatial templates. Our method is built on top of Deep Fragments \cite{karpathy2014deep}.
R@k is Recall@K (high is good), Mean r is the mean rank (low is good).
}
\label{table:pascal1k}
\end{table}
Our estimates of the spatial templates from the previous sections rely on the restricted language in form of structured queries 
and annotated bounding boxes. We now turn to the Pascal1k dataset that features natural language sentences and therefore requires us to deal with implicit supervision for learning representations of spatial relations. We improve over Deep Fragment Embeddings \cite{karpathy2014deep}\footnote{We downloaded the source code from \url{http://cs.stanford.edu/people/karpathy/defrag/code.zip}. 
Our performance numbers are on average slightly better then the reported ones in \cite{karpathy2014deep}, as the code has been improved after the publication.} to include our spatial model as discussed in section \ref{section:deep_fragments}.
For our method, we choose the dimension of a space of spatial concepts (section \ref{section:deep_fragments}) to be $4$, and a spatial representation of 20 pooling regions (precisely the 2-by-2 + 4-by-4 scheme) based on the validation set. Here, we treat a space of spatial concepts more abstractly and we do not associate the prepositions with the indices to spatial templates. Our spatial fragments are pairs of detections, and spatio-textual fragments are arbitrary triplets ($R$, $\bs{t_1}$, $\bs{t_2}$) from the dependency parser. We find it more effective to start the training with the only original model and next proceed to a joint training with our spatial extension. Following \cite{karpathy2014deep} we also compare our method against other embedding models on this dataset. Table \ref{table:pascal1k} shows that our model improves over Deep Fragment Embeddings and consistently outperforms other methods on both tasks: image retrieval and image annotation (here the method retrieves sentences based on the image). Adding our spatial model to Deep Fragment Embeddings improves R@10 by $1.4$ and $2.0$ units on 
both
tasks respectively. We have also implemented spatial model based on the distance and containment features \cite{golland2010game} but we didn't achieve satisfactory results - the model barely outperforms Deep Fragment Embeddings. Table \ref{table:pascal1k} proves the point that the state-of-the-art retrieval architectures benefit from a spatial model that we propose.
\\
\noindent{{\bf Improved and interpretable alignment}}
\begin{figure*}[t]
\centerline{\includegraphics[width=0.9\linewidth]{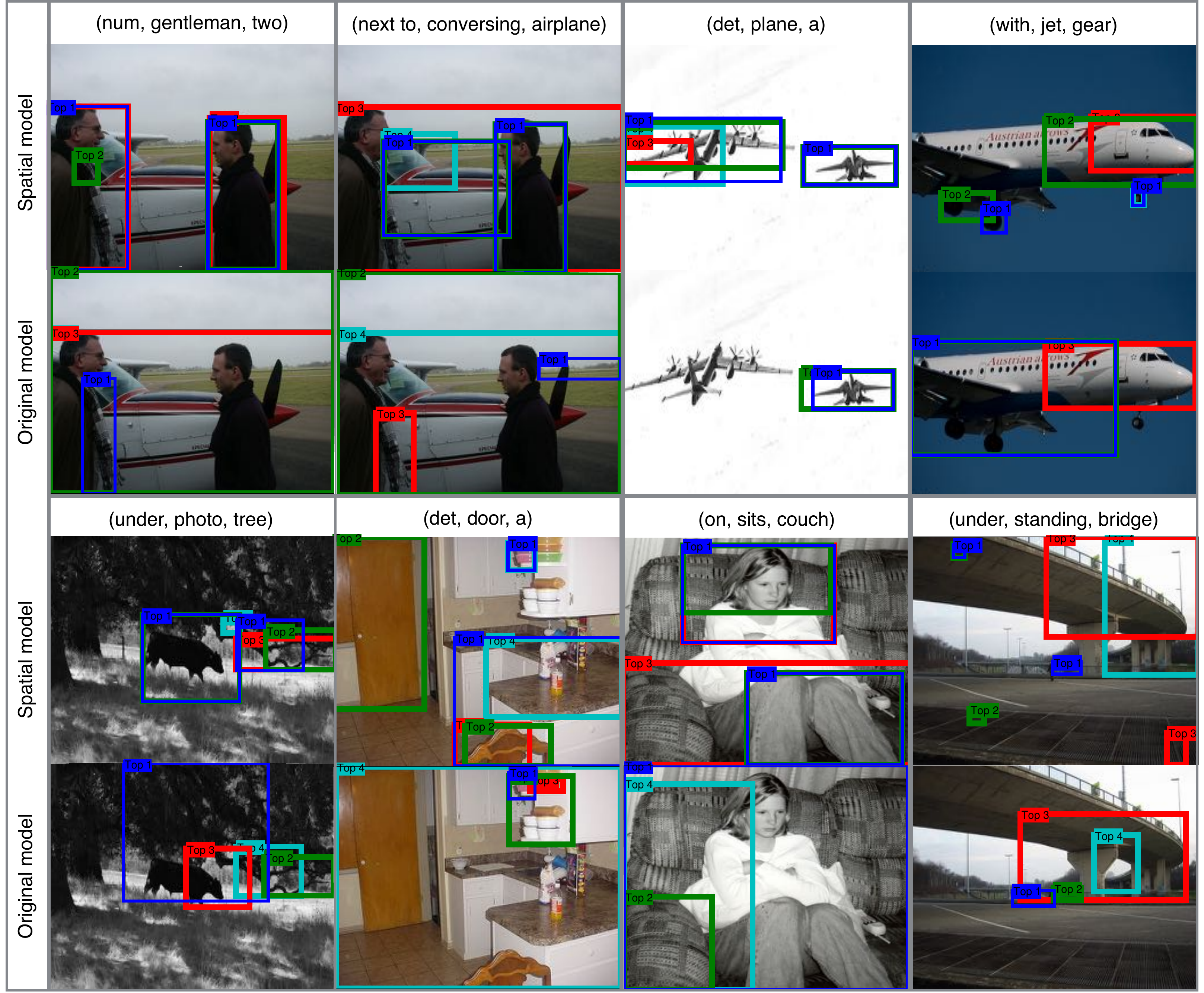}}
\caption{
Top $4$ best bindings between a textual fragment and all detections. Every column represents different textual fragments. The first and third rows show a spatial embedding. The second and fourth rows show an original embedding \cite{karpathy2014deep}. Colors encode scores of fragments associations. Starting from the top scoring: blue, green, red, and cyan. If two fragments overlap, we only show the top scoring one. Since spatial fragments represent pairs of detections, we use the same color encoding for the same pair. Best viewed in color.
}
\label{fig:localization_results}
\end{figure*}
Given a set of detections representing visual fragments and two words under a dependency relation representing textual fragments, Deep Fragment Embeddings learns a binding so that the dot product between the matching fragments is high.
Hence, for a textual fragment (dependency relation, word $1$, word $2$), we compute the scores between every detection and the textual fragment, and visualize top $4$ scoring bindings. 
As we argue in this work, the notion of fragments can naturally be generalized to pairs of detections that are in a spatial relation. This  is particularly attractive because of the symmetry to textual fragments that always take two words under some relation dependency into account. Fig. \ref{fig:localization_results} shows how alignment improves over the original non-spatial model. 

As an example the fragment ('num', 'gentleman', 'two'), which comes from a sentence 'Two gentleman talking in front of propeller plane', aligns well with a spatial fragment representing human detections. Another interesting example includes the fragment ('with', 'jet', 'gear') with the second top fragment that relates the plane's cockpit with its gears (the top scoring one relates two gears together). 
Such interpretability is often missing in the output of the original model (second and fourth rows of Fig. \ref{fig:localization_results}).

\section{Conclusion}
\label{section:conclusion}
We address the problem of missing spatial relations in modern retrieval architectures.  
Although the research on spatial concepts has a long tradition, it mostly concerns robotics. 
Even then, previous works use either rule-based approaches or a hand-crafted set of features. 
In contrast, our work links spatial models with spatial pooling regions framework and offer a simple and uniform framework for spatial reasoning. Next, we conduct several experiments where we show that a competitive pooling-based spatial model can be learnt solely from data. 
Our analysis on newly collected data shows that automatically generated queries from the previous work
have different distribution of spatial concepts than the real data. Moreover, our visualization of alignments suggests that spatial model improves bindings between fragments.
Finally, we hope that our results together with our data of spatial queries will foster further research on spatial concepts. 
For this purpose we will make our dataset publicly available.
In particular, we are excited to study other spatial categories and higher order spatial terms.

{\bf Acknowledgement:} We would like to thank Andrej Karpathy for his support and making his code available.

{
\small
\bibliographystyle{ieee}
\bibliography{egbib}
}

\end{document}